%% file: tr.tex
\documentclass[reqno]{amsart}
\usepackage{graphicx}
\usepackage{amsmath,amssymb,amsthm} 
\usepackage{bm}
\usepackage{color}
\usepackage[scriptsize,tight]{subfigure}
\usepackage{mathabx}
\usepackage[font=small]{caption}
\usepackage{ctable}
\usepackage[]{hyperref}

\DeclareMathOperator{\SE}{SE}

\DeclareMathOperator{\SO}{SO}

\DeclareMathOperator{\divv}{div} %

\newcommand{\otoprule}{\midrule[\heavyrulewidth]}
\DeclareMathOperator{\GL}{GL}
\DeclareMathOperator{\const}{const}

\title[Volumetric Reconstruction Applied to Studies of Size and Weight]{Volumetric Reconstruction Applied to Perceptual Studies of Size and Weight}
\author{J. Balzer, M. Peters, and S. Soatto}
\begin{document}

\maketitle

\begin{abstract}

We explore the application of volumetric reconstruction from structured-light sensors in cognitive neuroscience, specifically in the quantification of the \emph{size-weight illusion}, whereby humans tend to systematically perceive smaller objects as heavier. We investigate the performance of two commercial structured-light scanning systems in comparison to one we developed specifically for this application. Our method has two main distinct features: First, it only samples a sparse series of viewpoints, unlike other systems such as the Kinect Fusion. Second, instead of building a distance field for the purpose of points-to-surface conversion directly, we pursue a first-order approach: the distance function is recovered from its gradient by a \emph{screened Poisson reconstruction}, which is very resilient to noise and yet preserves high-frequency signal components. Our experiments show that the quality of metric reconstruction from structured light sensors is subject to systematic biases, and highlights the factors that influence it. Our main performance index rates estimates of volume (a proxy of size), for which we review a well-known formula applicable to incomplete meshes. Our code and data will be made publicly available upon completion of the anonymous review process. 

\end{abstract}

\section{Introduction}\label{sec:introduction}

\begin{figure}[tb]
\centering
\subfigure[]{\label{fig:introa}\includegraphics[height=2.4cm]{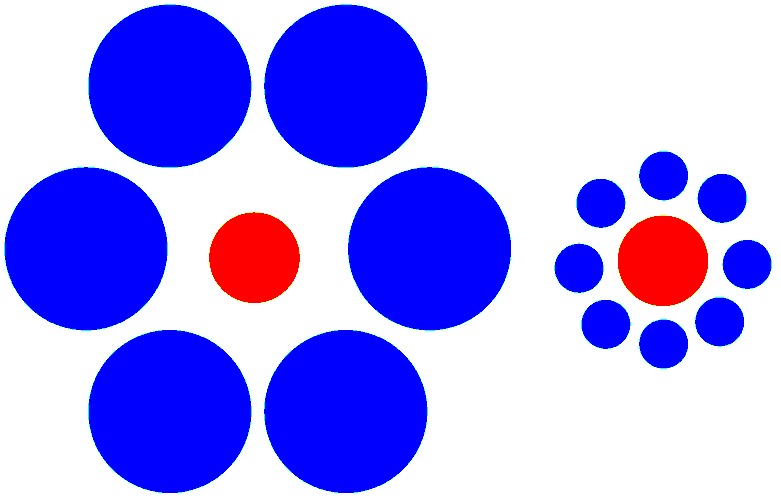}}\hspace{0.9cm}
\subfigure[]{\label{fig:introb}\includegraphics[height=2.4cm]{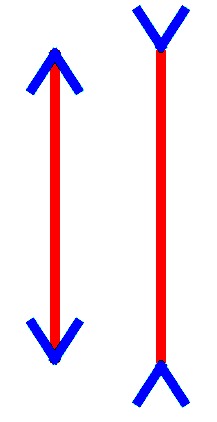}}\hspace{0.9cm}
\subfigure[]{\label{fig:introc}\includegraphics[height=2.4cm]{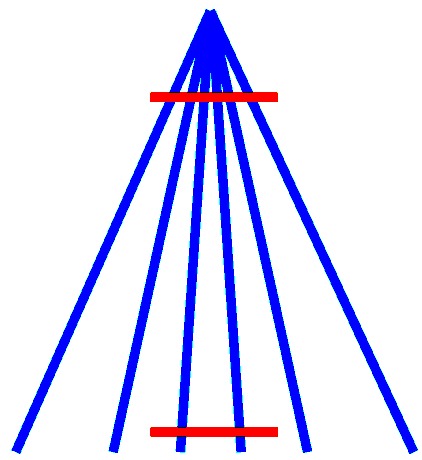}}\\
\subfigure[]{\centering\label{fig:mcubes}\includegraphics[width=0.8\columnwidth]{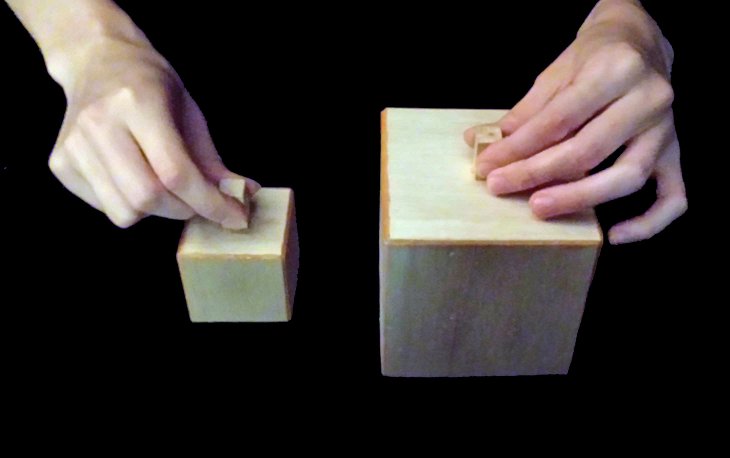}}
\caption{The \subref{fig:introa} Ebbinghaus, \subref{fig:introb}~M\"{u}ller-Lyer, and \subref{fig:introc}~Ponzo illusions demonstrate that perception is not always veridical \cite{Goodale11}. For each, the orange elements are identical, although they appear unequal. \subref{fig:mcubes} In the \emph{size-weight illusion}, two objects that weigh the same feel as if the smaller one is heavier \cite{Charpentier91}.}
\label{fig:intro}
\end{figure}

\subsection{Motivation}

We like to believe that our sensory systems provide us with precise and accurate information about objects within the environment, but our perception is often subject to systematic errors, or {\em illusions} (Figure~\ref{fig:intro}).  These can also occur between sensory modalities, often with visual information influencing haptic (touch) estimates of properties such as size or weight.  For example, a curious experience occurs when we lift two objects of equal weight but different size; systematically and repeatably, the smaller object feels heavier than the larger.  This \emph{size-weight illusion} (SWI) \cite{Charpentier91} cannot be explained by simple motor force error (i.e., it is not simply due to the production of more lifting or grip force for the larger object)~\cite{Flanagan00,Grandy06}, and so carries important implications for the dynamics of sensory integration between vision and haptics.  Likewise, altered visual appearance of an object (e.g. through stereoscopic goggles~\cite{Ernst02} or optical distortion with prisms~\cite{Rock64}) can significantly impact haptically-judged estimates of its size.  Simply put, when an object looks bigger than it really is, it feels bigger, too -- and any mismatch between vision and touch often goes completely unnoticed. 

In order to establish a solid quantitative empirical assessment of these illusions, we have developed methodologies to examine the relationship between true size and perceived size.  Previous investigations have uncovered evidence that the relationship between an object's true volume and its perceived volume often follows a power function with an average exponent of $0.704$ ($\sigma=0.08$)~\cite{Frayman81}.  However, these prior investigations have predominantly used objects which are geometric, symmetrical, and convex -- properties which alone cannot adequately capture the range of objects regularly encountered in everyday environments.  Thus, to systematically and comprehensively explore the relationship between true volume and perceived volume so as to better understand this percept's contribution to visual-haptic integration and, consequently, perception in general, we have developed dedicated methods to capture an ecologically valid set of stimuli.

\begin{figure}
\includegraphics[height=4cm]{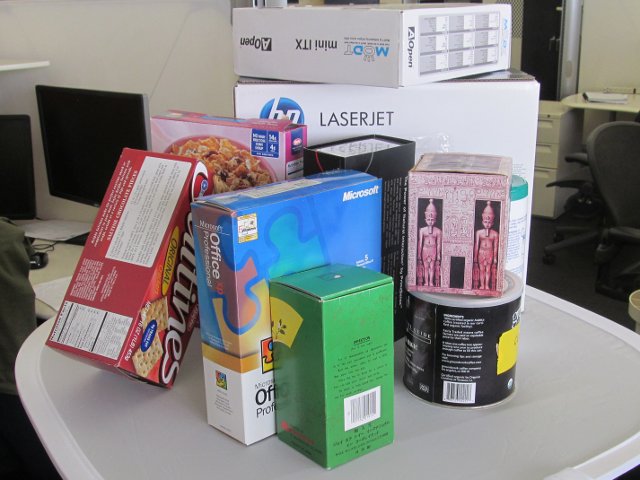}\hspace{0.8cm}
\includegraphics[height=4cm]{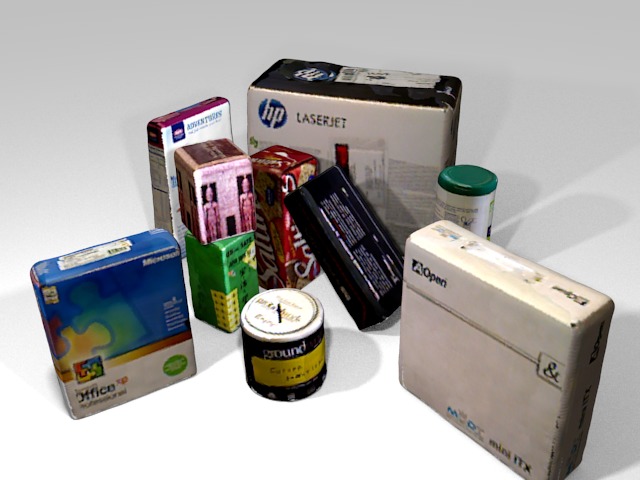}
\caption{We compare the performance of two scanning systems with the one proposed here at hand of a set of geometric primitives, whose volume can be easily measured by hand. Although shown as a collection, all items above have been reconstructed and texture-mapped individually.}\label{fig:cubes}
\end{figure}

Our goal is to build a data base of digital models of our specimens, which would allow us to infer volume and any other desired geometric properties. A method to create such models should meet the following list of criteria:
\begin{itemize}
\item It is mandatory that the sensing modality of choice be contactless: Cell phones and other hand-held consumer electronics, e.g., are among the classes of objects relevant to our psychological studies. They cannot simply be sunk in a fluid leveraging Archimedes' principle, neither can anything which is permeated by the fluid, as doing so would not provide the desired \emph{visible} volume. 
\item The underlying sensor should be  inexpensive and easy to use for non-experts; e.g., researchers in psychology and neuroscience. This rules out dedicated lab equipment, e.g., for white-light interferometry and the like. 
\item The resulting models should exhibit some topological structure: For volume computations, it must at least be possible to distinguish interior and exterior. Therefore, point clouds alone are insufficient in this regard. 
\item Given the complexity of everyday objects -- and that they frequently depart from the cubic, spherical, or cylindrical -- we target an improvement of accuracy over back-of-the-envelope estimates. 
\item A rich spectrum of object classes is covered in terms of admissible geometry and reflectance properties. Specular surfaces, e.g., would need to be coated with powder to make them amenable to laser scanning. But this would contravene the first criterion and thus eliminates laser scanning from the list of candidates.
\end{itemize}
In light of these requirements, we opt for triangulation based on structured-light encoding as the primary sensing modality. The principle behind this method has been known for decades but has seen a renaissance in computer vision ever since Primesense introduced a fully functional color-range (``RGBD'') sensor unit in integrated-circuit design. This system-on-chip later became the core component of Microsoft's Kinect, which subsequently had an considerable impact in geometry reconstruction, tracking, occlusion detection, and action recognition, among other applications. 

\subsection{Contribution and overview}

We develop a system for structured-light scanning of small- to medium-scale objects, which we dub \emph{Yet Another Scanner}, or YAS. Naturally, the question arises why we would need yet another scanner when several systems and commercial products are already available, e.g., Kinect Fusion~\cite{Newcombe2011},  ReconstructMe\footnote{\href{http://reconstructme.net/}{http://reconstructme.net/}}, Artec Studio\footnote{\href{http://www.artec3d.com/software/}{http://www.artec3d.com/software/}}, KScan3D\footnote{\href{http://www.kscan3d.com/}{http://www.kscan3d.com/}}, Scanect\footnote{\href{http://skanect.manctl.com/}{http://skanect.manctl.com/}}, Scenect\footnote{\href{http://www.faro.com/scenect/}{http://www.faro.com/scenect/}}, and Fablitec's 3d scanner\footnote{\href{http://www.fablitec.com/}{http://www.fablitec.com/}}; in particular, when most of these generate visually highly-pleasing results. 

The main reason is that, albeit visually pleasing, the reconstructions provided by these methods are subject to biases that make them unsuitable for scientific investigation. The analysis, which is presented in Sect.~\ref{subsec:results}, compares the performance of YAS with that of two competing state-of-the-art implementations. While the reconstruction algorithm described in Sects.~\ref{subsec:alignment} and~\ref{subsec:poisson} itself is not novel, we carefully justify all choices to be made in its design w.r.t. above-listed requirements. Additionally, we address the issue that an aligned series of range images suffers from incompleteness precisely where the ground plane supports the object. Sect.~\ref{subsec:volumeestimation} proposes a strategy to circumvent this problem in volume estimation which avoids complicated hole-filling algorithms. We believe that a tool which outperforms commercial software but is accessible for further scientific development may be of interest to the computer vision community as well. Hence, as the final contribution, we will distribute the source through the repository at \href{https://bitbucket.org/jbalzer/yas}{https://bitbucket.org/jbalzer/yas}.

\section{System description}

\subsection{Data acquisition and calibration}

In all our experiments studies, we used Microsoft's Kinect and Primesense's Carmine 1.09. Both devices are shipped with a factory calibration of depth and RGB camera intrinsics as well as the coordinate transformation between their local reference frames. Initial visual assessment (by the naked human eye) approves of the default calibration simply because the point clouds computed from the range image seem to be accurately colored by the values of the RGB image. Extensive tests, however, have shown that -- in the spirit of our introductory remarks -- such an evaluation is misleading, and significant metric improvements through manual re-calibration are possible. For this purpose, we rely on the toolbox accompanying the paper~\cite{DanielHerreraC2012} to estimate all aforementioned parameters plus a depth uncertainty pattern, which varies both spatially and with depth itself.

\subsection{View alignment}\label{subsec:alignment}

\begin{figure}
\centering
\subfigure[Stratification]{\label{fig:misalign1}\includegraphics[height=4cm]{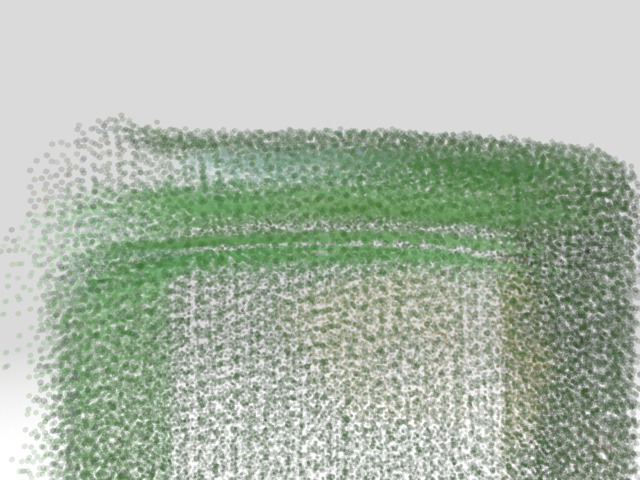}}\hspace{0.8cm}
\subfigure[Accumulation of tracking errors]{\label{fig:misalign2}\includegraphics[height=4cm]{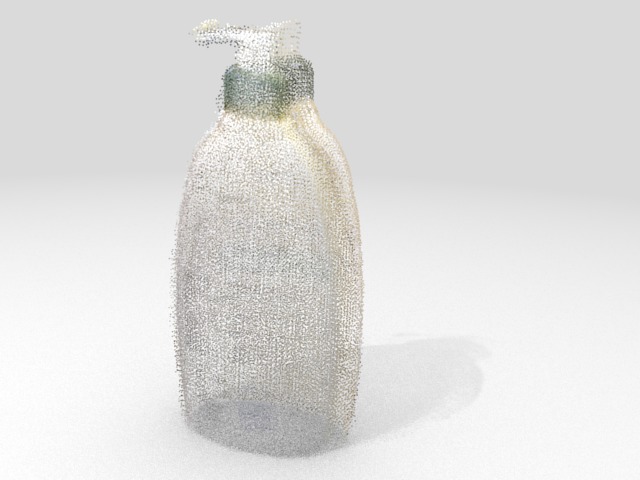}}\\
\subfigure[Scenect]{\label{fig:misalign3}\includegraphics[height=4cm]{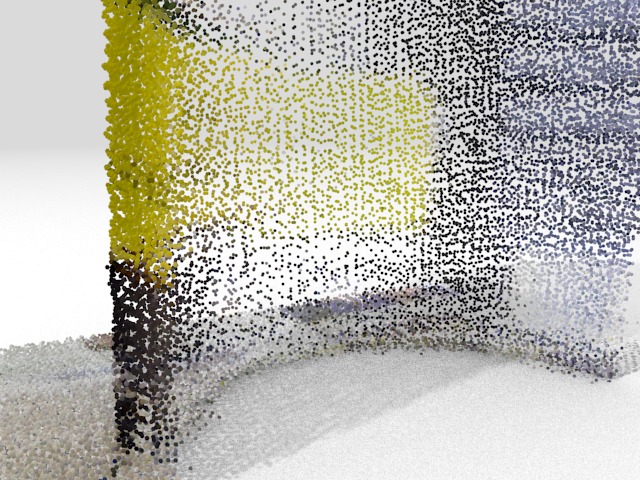}}\hspace{0.8cm}
\subfigure[YAS]{\label{fig:misalign4}\includegraphics[height=4cm]{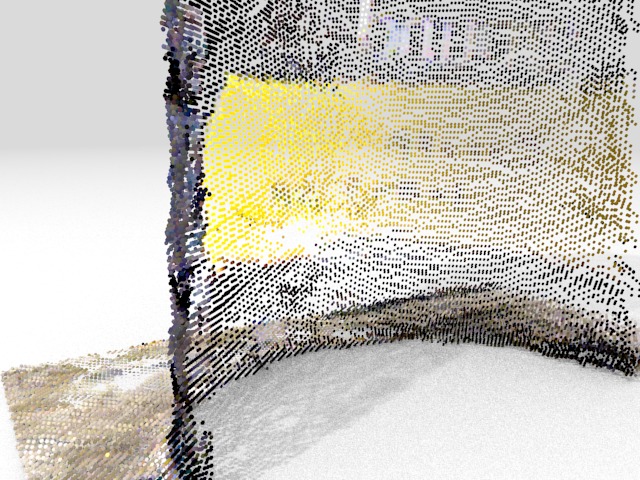}}
\caption{We promote a sparse sampling of viewpoints for covering the object of interest with depth map measurements, for \subref{fig:misalign1} this reduces the impact of stratification and \subref{fig:misalign2} reduces the probability of drift or tracking failure. \subref{fig:misalign3}-\subref{fig:misalign4} The difference in alignment quality only becomes noticeable in cross-sections of the point clouds.}\label{fig:stratification}
\end{figure}

A calibrated sensor immediately delivers physically plausible depth data. The integration of measurements from different vantage points into a common 3-d model can thus be seen as the core challenge here. A comprehensive overview of the state of the art in scan alignment is found in the recent survey~\cite{Tam2013}. Essentially, one can distinguish between two approaches: \emph{tracking} and \emph{wide-baseline matching}. The former is at the heart of Kinect Fusion~\cite{Newcombe2011} and the majority of commercially available software. Its main motivation stems from the fact that correspondence is easier to establish when two images haven been acquired closely in time -- supposing, of course, that the motion the camera has undergone between each image acquisition and the next meets certain continuity constraints. We believe, however, that for the purpose of small-scale object reconstruction, the disadvantages of tracking predominate. First and foremost, there is the question of redundancy: How do we deal 
with the stream of depth data when operating an RGBD camera at frame rates up to $30~\mathrm{fps}$? On the one hand, redundancy is desirable because single depth images may not cover the entire surface of the unknown object, e.g., due to occlusions or radiometric disturbances of the projected infrared pattern. On the other hand, integration of overcomplete range data into a common 3-d model puts high demands on the quality of alignment. Most feature trackers operate on a reduced motion model\footnote{E.g., the Lucas-Kanade method assumes pure translational motion.} and are thus prone to drift. Such deviations in combination with the uncertainty in the raw depth data can lead to a \emph{stratification} of points in regions appearing in more than a single image. This effect is illustrated in Fig.~\ref{fig:misalign1}, which becomes more severe with higher numbers of processed images. 

Kinect Fusion~\cite{Newcombe2011} deals with redundancy by instantaneously merging the depth stream into an implicit surface representation over a probabilistic voxel grid. The extra dimension, however, raises memory consumption -- even in implementations utilizing truncation or an efficient data structure such as an octree. Also, without a-priori knowledge of the specimen's size, it is difficult to gauge the interplay between the spatial resolutions of 3-d grid and raw depth data, the latter being left exploited only sub-optimally. Last but not least, a system based on tracking is little user-friendly: It requires the operator to move the sensor as steadily as possible. Otherwise, temporal under-sampling or motion blur can lead to a total breakdown of the alignment process.

Here, we closely follow the wide-baseline matching procedure developed by the robotics community, notably in the work of Henry et al.~\cite{Henry2012}. It follows two quasi canonical steps: First, a set of local descriptors around interest points in each RGBD image is computed as well as a set of tentative matches between them. Such descriptors incorporate radiance and depth information either exclusively or in combination. Second, subsets of cardinality three are selected from all matches at random. Each subset admits a hypothesis about the rigid motion that transforms one of the point clouds into the other. The winning hypothesis is taken to be the one supporting the most matches, i.e., generating the highest number of \emph{inliers}~\cite{Fischler1981}. We review these initial two steps in the following sections.

\subsubsection{Sampling}\label{subsubsec:sampling}

Let us formally consider the case of two views $l=0,1$, i.e., we look for two rigid motions $g_l\in\SE(3)$, $g_l:\bm{x}\mapsto\mathbf{R}_l\bm{x}+\bm{t}_l$, with $\mathbf{R}_l\in\SO(3)$ and $\bm{t}_l\in\mathbb{R}^3$. Without loss of generality, one can assume that $g_0$ coincides with the world reference frame, i.e., $\mathbf{R}_0=\mathbf{I}$ and $\bm{t}_0=\bm{0}$, which leads to a simpler notation of the unknowns $\mathbf{R}_1=\mathbf{R}$ and $\bm{t}_1=\bm{t}$.

A number of interest points $\{\bm{p}^i_0\}, \{\bm{p}^j_1\}$ with high response is extracted from each of the two RGB images corresponding to $g_{0,1}$ by means of the SIFT detector. The points are combined into a set of putative correspondences $\mathcal{C}=\{(\bm{p}^{i_k}_0,\bm{p}^{j_k}_1)\;|\; k=1\ldots,n\in\mathbb{N}\}$ by thresholded forward-backward comparison of the distances between associated SIFT descriptors~\cite{Lowe1999}. The search for nearest neighbors can be sped up by a locality-sensitive hashing technique or similar. However, we found in all of our experiments that the time consumed by a brute-force search was within acceptable limits. Next, we repeatedly draw a sample of three matches from $\mathcal{C}$ and obtain a set of triples $\mathcal{H}=\{(k_1,k_2,k_3)\;|\; 1\leq k_1,k_2,k_3 \leq n,k_1 \neq k_2 \neq k_3 \}$.

\subsubsection{Consensus}\label{subsubec:consensus}

Implicitly, each of the elements of $\mathcal{H}$ determines a hypothesis about the transformation we are looking for: Suppose we already know $g_1\in\SE(3)$, then the geometric least-squares error for some $(k_1,k_2,k_3)\in\mathcal{H}$ is given by
\begin{equation}\label{eq:lse}
	e(k_1,k_2,k_3)=\sum\limits_{k\in\{k_1,k_2,k_3\}}\frac{1}{2}\|\bm{x}^{i_{k}}_0-\mathbf{R}_k\bm{x}^{j_k}_1-\bm{t}_k\|^2.
\end{equation}
Here, the six points $\bm{x}_0^{i_{k}},\bm{x}_1^{j_{k}}\in\mathbb{R}^3$ equal the backprojections of the three matches $(\bm{p}^{i_k}_0,\bm{p}^{j_k}_1)$ forming the current hypothesis. Given the intrinsic camera parameters, they can be easily computed from the data delivered by the calibrated depth sensor. Conversely, given a triple $(k_1,k_2,k_3)\in\mathcal{H}$, we can find a global minimizer $g_1^*$ of the convex function~\eqref{eq:lse} in the following way: Denote by $\bar{\bm{x}}_0$ the mean of $\bm{x}_0^{i_{k}}$ over $k$ and define $\bar{\bm{x}}_0^{i_{k}}=\bm{x}_0^{i_{k}}-\bar{\bm{x}}_0$. The quantities $\bar{\bm{x}}_1$ and $\bar{\bm{x}}_1^{j_{k}}$ are defined analogously. A minimizer in the \emph{entire general linear group} of matrices $\GL(3)$ is
\[
	\mathbf{H}=\sum\limits_{k\in\{k_1,k_2,k_3\}}\bar{\bm{x}}_0^{i_{k}}(\bar{\bm{x}}_1^{j_{k}})^{\top},
\]
cf.~\cite{Williams2001}. One needs to make sure that the optimal $g_1^*$ involves a genuine \emph{rotation matrix} by projecting $\mathbf{H}$ onto $\SO(3)$. This is commonly achieved by \emph{Procrustes analysis}, essentially consisting of a singular-value decomposition: Write $\mathbf{H}$ as the product $\mathbf{U}\mathbf{\Sigma}\mathbf{V}^{\top}$ with two orthogonal factors $\mathbf{U},\mathbf{V}\in\mathbb{R}^{3\times 3}$, then $\mathbf{R}_k^*=\mathbf{V}\mathbf{U}^{\top}$. Once $\mathbf{R}_k^*$ is known, the optimal translation vector can be computed as $\bm{t}_k^*=\bar{\bm{x}}_0-\mathbf{R}_k^*\bar{\bm{x}}_1$. The transformation for the element of $\mathcal{H}$ attaining the highest consensus among \emph{all} matches in $\mathcal{C}$ constitutes the solution to the global alignment problem~\cite{Fischler1981}. The result is refined based on the inlier correspondences with the iterative closest-point (ICP) method~\cite{Besl1992}. This also ensures a \emph{geometrically} continuous alignment, which is not guaranteed because $\mathcal{H}$ was generated 
merely based on \emph{photometry}.

\subsection{Surface reconstruction}\label{subsec:poisson}

The common point cloud obtained after merging all aligned depth maps carries no information about the topological relationship between its elements, but as we will see shortly, such information plays a crucial part in volume estimation. There exists a wealth of algorithms for point-to-surface conversion, most of which depend on the signed or unsigned Euclidean distance field $\varphi:\mathbb{R}^3\to\mathbb{R}$ induced by the point cloud. Kinect Fusion, e.g., computes $\varphi$ directly. Alternatively, when the point cloud is oriented, i.e., each point $\bm{x}$ is endowed with an estimate of the normal vector $\bm{n}$ the surface should have at that location, one can search for the function $\varphi$ minimizing
\begin{equation}\label{eq:dirichlet}
	\int\limits_D\frac{1}{2}\|\nabla\varphi-\bm{n}\|^2 \mathrm{d}\bm{x}\rightarrow\min.
\end{equation}
Here, with slight abuse of notation, $\bm{n}$ refers to an (arbitrary) continuation of the normal field from the point set to some sufficiently large rectangular domain $D\subset\mathbb{R}^3$. Since the gradient of any scalar function is orthogonal to its level sets, this gives a family of \emph{integral surfaces} 
\begin{equation}\label{eq:isosurface}
\Gamma = \{\bm{x}\in\mathbb{R}^3\,|\,\varphi(\bm{x})=C\}
\end{equation}
of $\bm{n}$. A minimizer of~\eqref{eq:poisson} is found as the solution of the Euler-Lagrange equation
\begin{equation}\label{eq:poisson}
	\Delta \varphi = \divv\bm{n}
\end{equation}
under natural boundary conditions (here of Neumann type). Eq.~\eqref{eq:poisson} is the well-known Poisson equation and eponymous for the \emph{Poisson reconstruction} algorithm proposed in~\cite{Kazhdan2006}. 

The motivation for increasing the order of differentiation as compared to the direct approach (i.e., that followed in Kinect Fusion) is twofold: First, Eq.~\eqref{eq:dirichlet} is a variant of the Dirichlet energy, which implies that small holes in the point cloud, i.e., areas where $\bm{n}=\bm{0}$, will automatically be in-painted harmonically. Second, for a solution to exist in the strong sense $\nabla\varphi=\bm{n}$, the normal field must be \emph{integrable} or \emph{curl-free}. Noise, which is very common in RGBD images, is responsible for most of the non-integrability in a measured normal field $\bm{n}$. In the variational setting~\eqref{eq:dirichlet}, however, $\bm{n}$ is implicitly replaced by the next-best gradient (its so-called \emph{Hodge projection}, cf.~\cite{Cantarella2002}), which makes the approach very resilient to stochastic disturbances but at the same time destroys fine details. The smoothing effect can be mitigated by imposing Dirichlet conditions on~\eqref{eq:poisson} at a sparse set of salient points. This so-called \emph{screened Poisson reconstruction} has recently been introduced in~\cite{Kazhdan2013}. 

The point cloud is easily oriented exploiting the known topological structure of the pixel lattice: Given a depth parametrization of the surface $z(x,y)$ over the two orthogonal camera coordinate directions $x$ and $y$, the normal can be written as $\bm{n}=(-\partial_x z, -\partial_y z,1)^{\top}$. The partial derivatives of $z$ w.r.t. camera and image coordinates $(x,y)^{\top}$ respectively $(u,v)^{\top}$ are related by the chain rule:
\begin{equation}\label{eq:normal}
	\frac{\partial z}{\partial x} = \frac{\partial z}{\partial u}\frac{\partial u}{\partial x}=\frac{f_u}{z}\frac{\partial z}{\partial u},\quad	\frac{\partial z}{\partial y} = \frac{\partial z}{\partial v}\frac{\partial v}{\partial y}=\frac{f_v}{z}\frac{\partial z}{\partial v}.
\end{equation}
Here, $f_u, f_v$ are the focal lengths of the pinhole depth camera, and finite-differencing provides an approximation to the gradient of $z(u,v)$. For the details of numerically solving~\eqref{eq:poisson} and selecting the constant $C$ in~\eqref{eq:isosurface} appropriately, we refer the reader to the original paper~\cite{Kazhdan2006}. 

\subsection{Volume estimation}\label{subsec:volumeestimation}

\subsubsection{Closed surfaces}\label{subsubsec:volcont}

Suppose for the moment that $\Gamma$ given by~\eqref{eq:isosurface} is compact and closed. The volume $V$ of the domain $\Omega\subset\mathbb{R}^3$ it encompasses is defined as the integral of the characteristic function $\chi_{\Omega}$ of $\Omega$:
\begin{equation}\label{eq:volume}
	V=|\Omega|=\int\limits_{\mathbb{R}^3}\chi_{\Omega}\mathrm{d}\Omega=\int\limits_{\Omega}1 \mathrm{d}\Omega.	
\end{equation}
Unfortunately, an evaluation of this integral is not very practical for two reasons.  First, doing so would require a regular grid over $\Omega$, which introduces undesirable artefacts where it interacts with a discrete version of $\Gamma$. Second, an expensive nearest-neighbor problem would need to be solved\footnote{But could be remedied by re-visiting the signed distance function $\varphi$ of $\Gamma$ from Sect.~\ref{subsec:poisson}.} to determine whether a point is inside or outside of $\Omega$. The following trick is based on the classic Gauss divergence theorem, cf.~\cite{Mirtich1996}, which relates the flow of \emph{any} continuously differentiable vector field $\bm{v}:\mathbb{R}^3\to\mathbb{R}^3$ through the boundary $\Gamma=\partial\Omega$ of $\Omega$ with its source density or \emph{divergence} in the interior:
\begin{equation}\label{eq:gauss}
	\int\limits_{\Omega}\divv\bm{v}\mathrm{d}\Omega=\int\limits_{\Gamma}\langle\bm{v},\bm{n}\rangle \mathrm{d}\Gamma.
\end{equation}
The left-hand side of this equation does not quite resemble the right-hand side of~\eqref{eq:volume}, yet. However, this can be achieved by a clever choice of $\bm{v}$, e.g., $\bm{v}:=(x,0,0)^{\top}$, but note that several variants will work equally well and that $\bm{v}$ is \emph{not} unitary. We will return to this point later in Sect.~\ref{subsubsec:holefilling}. We have $\divv \bm{v}=1$ so that combining~\eqref{eq:volume} and~\eqref{eq:gauss} yields
\begin{equation}\label{eq:volintegral}
	V=\int\limits_{\Gamma}\langle\bm{v},\bm{n}\rangle \mathrm{d}\Gamma.
\end{equation} 

\begin{figure}[tb]
\centering
\def\svgwidth{0.8\columnwidth}
{\scriptsize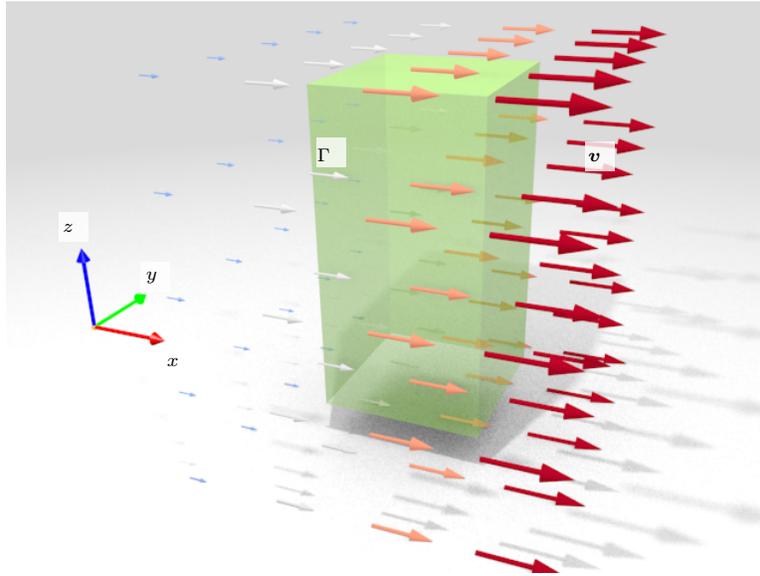}
\caption{The volume of a compact object $\Omega$ equals the mean flow of the vector field $\bm{v}=(x,0,0)$ through its boundary surface $\Gamma$. If the support of the object is aligned with the $xy$- or $xz$-plane, it is not traversed by any stream line, and hence needs not be explicitly filled with mesh faces for a faithful volume estimate. The flow field is colored by increasing magnitude $\|\bm{v}\|$.}\label{fig:vol}
\end{figure}

Let us look at the discrete case: Here, a level set $\Gamma_h$ of~\eqref{eq:isosurface} is extracted by the marching cubes in the form of a triangular mesh~\cite{Lorensen1987}. Such a piece-wise linear representation of the geometry provides a likewise locally-linear approximation of any function $f$ whose values $f_i$ are known at the vertices $\bm{x}_i\in\Gamma_h$, $i\in\mathbb{N}$. A Gauss-Legendre quadrature rule for $f$ with linear precision defined over the triangle $T$ is
\[
	\int\limits_{T}f\mathrm{d}\bm{x}\approx A(T)\sum_{i_k\in\mathcal{I}(T)}^3 \frac{1}{3}f_{i_k},
\]
where $A(T)$ equals the area of $T$, and $\mathcal{I}(T)$ enumerates its three corner vertices. Now substitute $f_i$ by the flow $\langle\bm{v}_i,\bm{n}_i\rangle$. The vertex normals $\bm{n}_i$ are usually taken to be the normalized mean of the face normals in a one-ring neighborhood of $\bm{x}_i$. Altogether, we finally obtain the following approximation of Eq.~\eqref{eq:volintegral}
\[
	V\approx\sum_{T\in \Gamma_h}\frac{A(T)}{3}\sum_{i_k\in\mathcal{I}(T)}^3 \langle\bm{v}_{i_k},\bm{n}_{i_k}\rangle.
\]

\subsubsection{Surfaces with boundary}\label{subsubsec:holefilling}

As explained in Sect.~\ref{subsec:poisson}, Poisson reconstruction accounts for most smaller holes in the aligned point clouds. The support of the object, i.e., its ``bottom'' or the area where it is in contact with the ground plane, however, remains usually unfilled. At the beginning of Sect.~\ref{subsubsec:volcont}, we demanded that $\Gamma$ be compact and closed because only then the volume of $\Omega$ is well-defined. We can lift this assumption in parts simply by a coordinate transformation: Remember that we chose $\bm{v}$ to point in the direction of the $x$-axis of the world coordinate system. Consequently, the flow through any of the planes $y=\const$ or $z=\const$ vanishes. As shown in Fig.~\ref{fig:vol}, all we have to do is align the support of the model with one of these planes. Without loss of generality, we choose $\{(x,y,z)^{\top}\in\mathbb{R}^3\,|\, z=0 \}$. In our scanning scenario, it is reasonable to assume that the specimens to be measured are spatially isolated enough that the depth images capture a 
significant portion of the ground plane surrounding the object, which can thus be detected fully automatically. To this end, we again invoke a RANSAC-type procedure, which samples triplets of points, calculates their common plane as a putative solution, and evaluates each such hypothetical plane by how many other points in the cloud it contains. 

\section{Experiments}\label{sec:experiments}

\begin{figure}
\subfigure[YAS]{\label{fig:rec_vs_gt}\includegraphics[height=3cm]{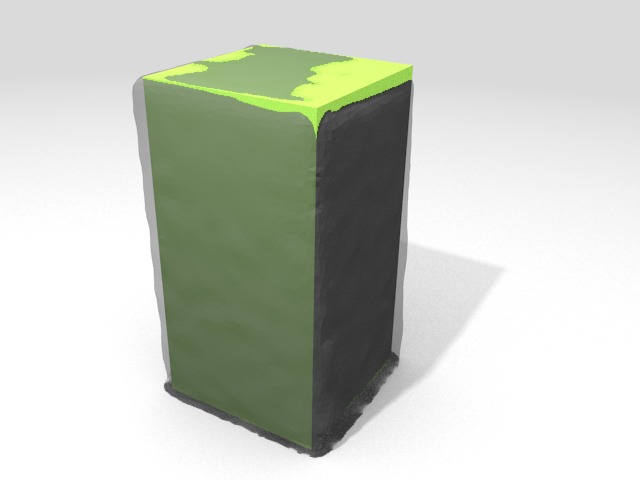}}\hfill
\subfigure[Scenect]{\label{fig:rec_vs_gt_scenect}\includegraphics[height=3cm]{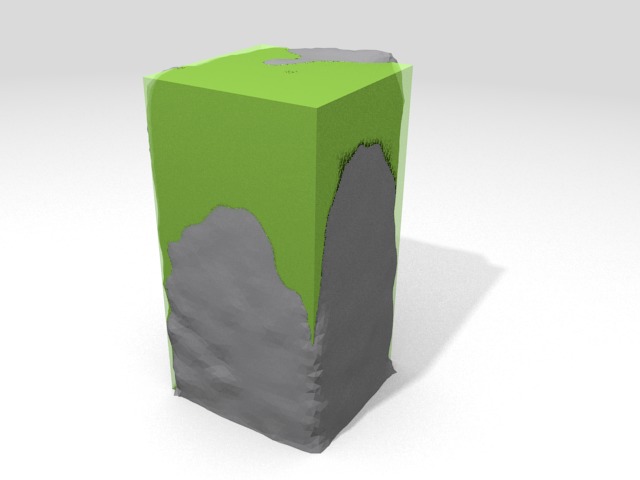}}\hfill
\subfigure[Kinect Fusion]{\label{fig:rec_vs_gt_kinfu}\includegraphics[height=3cm]{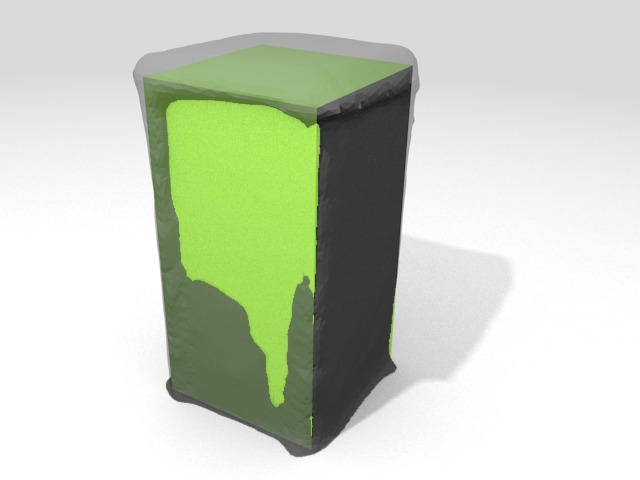}}
\caption{Reconstructions of item no. $1$ in the cube data set. The ground truth is shown in green. It becomes apparent that Kinect Fusion Explorer systematically overestimates the object volume.}\label{fig:rec_vs_gt}
\end{figure}

\subsection{Implementation}\label{subsec:implementation}

We created a C++ implementation of most of the reconstruction pipeline, including raw data acquisition, coarse and fine registration as well as detection of the ground plane. Ease of use is of premier priority in view of the interdisciplinary nature of this project. Therefore, the number of dependencies was kept as small as possible: The OpenCV library supplies us with all functionality for feature matching (Sect.~\ref{subsubsec:sampling}). Our implementation of the ICP method requires fast nearest-neighbor lookup which is based on the kd-tree data structure from the ANN library. We also created a graphical QT frontend which is showcased in the video included in the supplemental material. Poisson reconstruction (Sect.~\ref{subsec:poisson}) is currently done in Meshlab but will be integrated into our code in upcoming releases. 

We compare our method to the Kinect Fusion algorithm and Scenect, which is one of the few commercial software packages without hindering functionality restraints in the trial version. To warrant a fair comparison, all participating systems should be operated with the same sensor and the same intrinsic calibration. This proved to be somewhat difficult: Both Scenect and YAS access devices through the OpenNI framework driver supporting all of Primesense's products and the Xtion by Asus among others. The Point Cloud Library provides an open-source version of the Kinect Fusion algorithm which could potentially function with the Carmine 1.09 as well, but our experiences with it were little encouraging. For this reason, we had to resort to Microsoft's own implementation Kinect Fusion Explorer, which works exclusively with proprietary hardware, the Kinect. 

\subsection{Results}\label{subsec:results}

\begin{table}[tbp]
\centering
 \setlength{\tabcolsep}{1.5pt}
\def \mycolwidth{0.55cm}
\tiny
\begin{tabular}{lccccccc}\toprule
No. & $V_{\mathrm{manual}}\atop [10^{-3}m^3]$ & $V_{\mathrm{YAS}}\atop [10^{-3} m^3]$ & $E_{\mathrm{YAS}}\atop[\%]$ & $V_{\mathrm{Scenect}}\atop[10^{-3}m^3]$ & $E_{\mathrm{Scenect}}\atop[\%]$ & $V_{\mathrm{KinFu}}\atop[10^{-3}m^3]$ & $E_{\mathrm{KinFu}}\atop[\%]$ \\\otoprule %
1 & $1.26$ & $1.26$ & $0.11$ & $1.26$ & $0.6$ & $1.47$ & $16.6$ \\\midrule
2 & $2.82$ & $2.72$ & $-3.38$ & $2.68$ & $-5.1$ & $3.61$ & $28.4$\\\midrule
3 & $1.29$ & $1.21$ & $-6.07$ & $1.21$ & 	$-6.17$ & $1.63$ & $27.0$\\\midrule
4 & $3.07$ & $3.07$ & $-0.02$ & $2.83$ & $-7.57$ & $4.23$ & $38.0$\\\midrule
5 & $2.18$ & $2.06$ & $-5.44$ & $2.1$ & $-3.79$ & $2.85$ & $31.1$\\\midrule
6 & $6.26$ & $6.44$ & $2.86$ & $5.76$ & $-7.07$ & $7.69$ & $22.9$\\\midrule
7 & $1.66$ & $1.79$ & $8.22$ & $1.75$ & $5.65$ & $1.67$ & $0.55$\\\midrule
8 & $1.78$ & $1.89$ & $6.0$ & $1.67$ & $-6.28$ & $2.15$ & $20.3$\\\midrule
9 & $2.75$ & $2.75$ & $0.05$ & $2.68$ & $2.41$ & $3.84$ & $39.8$\\\midrule
10 & $24.6$ & $24.8$ & $1.05$ & $18.6$ & $-24.4$ & $28.8$ & $17.1$\\\bottomrule
$\mu$ &  &  & $-0.34$ &  & $-5.74$ & & $24.2$\\\midrule
$\sigma$ &  &  & $4.59$ &  & $7.76$ & & $11.5$ \\\bottomrule
\end{tabular}
\caption{Volume estimates and their relative error w.r.t. the value obtained by manual measurement.}
\label{tab:volumes}
\end{table}

\begin{figure}[tbp]
\subfigure[Volume]{\label{fig:volumes}\includegraphics[width=0.495\columnwidth]{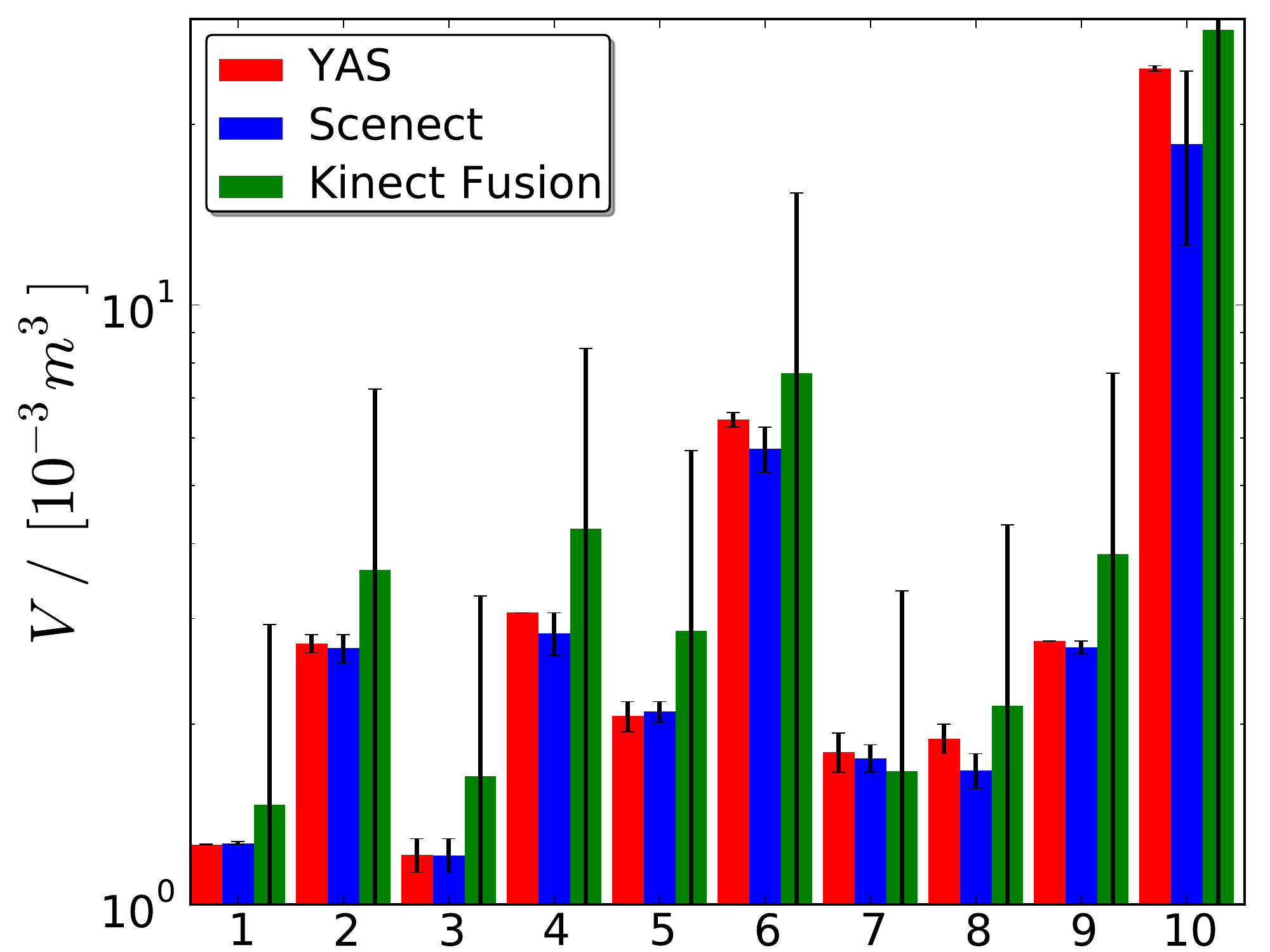}}\hfill
\subfigure[Error]{\label{fig:errors}\includegraphics[width=0.495\columnwidth]{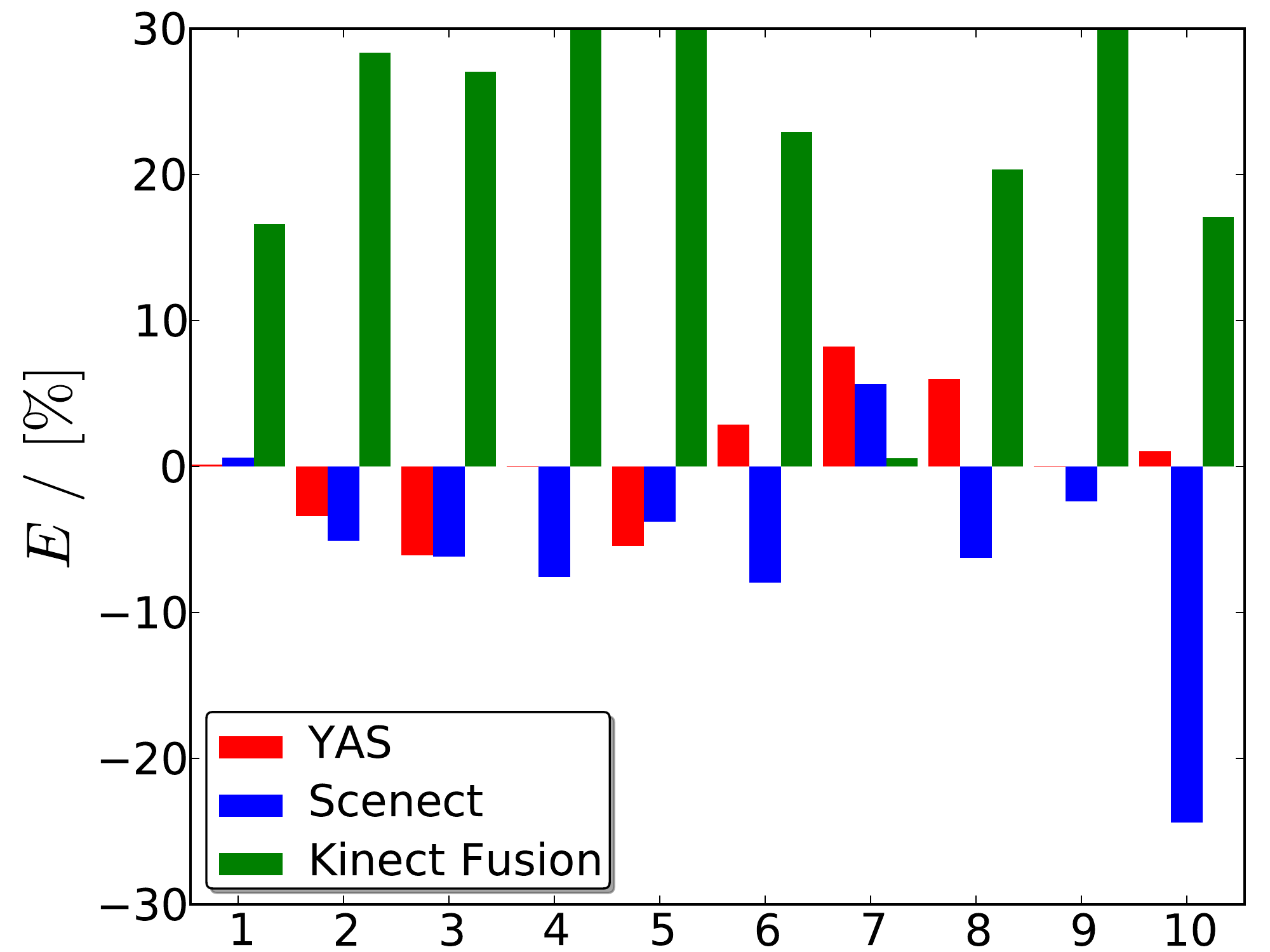}}
\caption{Visualization of Tab.~\ref{tab:volumes}. The error bars in~\subref{fig:volumes} indicate the absolute deviation from the ground truth volume. From~\subref{fig:errors}, it seems like Kinect Fusion is biased towards excessive values.}\label{fig:plots}
\end{figure}

\begin{figure}[tbp]
\centering
\setlength{\tabcolsep}{0pt}
\def \myfigheight{2cm}
\subfigure[Object]{\label{fig:photos}
\begin{tabular}[b]{c}
\includegraphics[height=\myfigheight]{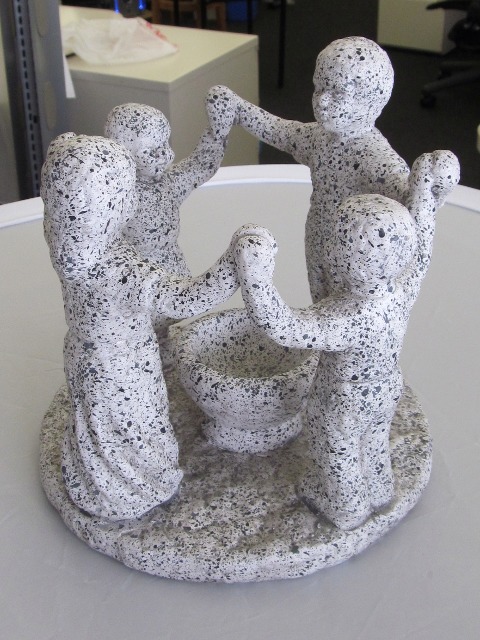}\\
\includegraphics[height=\myfigheight]{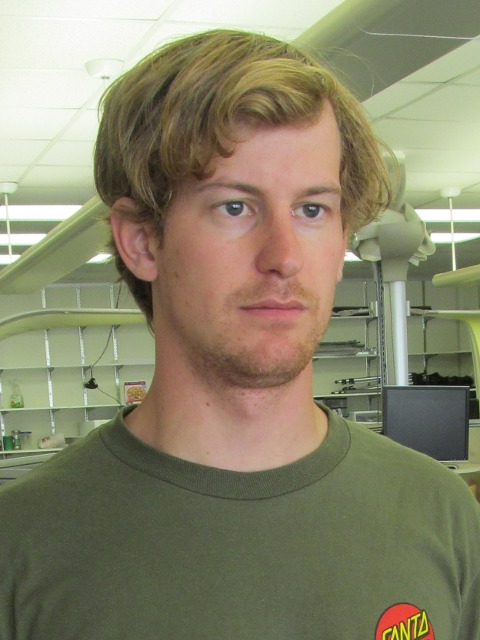}\\
\includegraphics[height=\myfigheight]{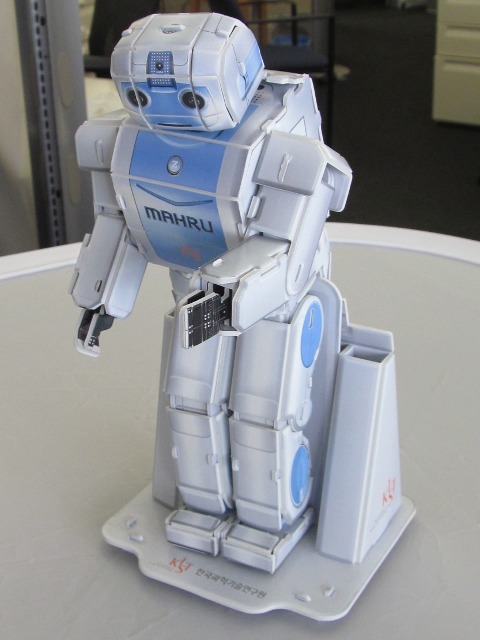}\\
\includegraphics[height=\myfigheight]{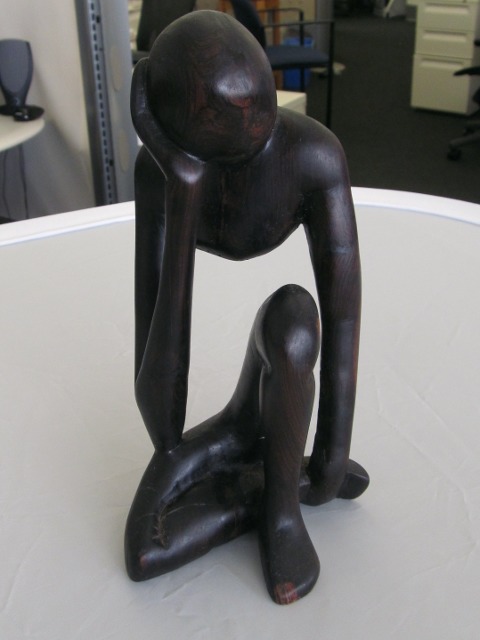}\\
\includegraphics[height=\myfigheight]{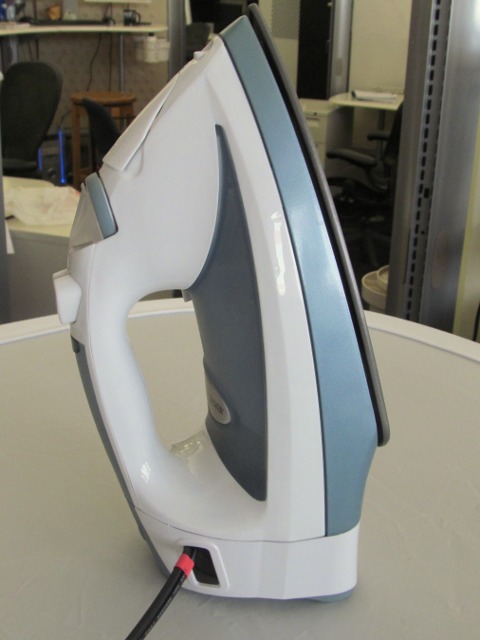}\\
\includegraphics[height=\myfigheight]{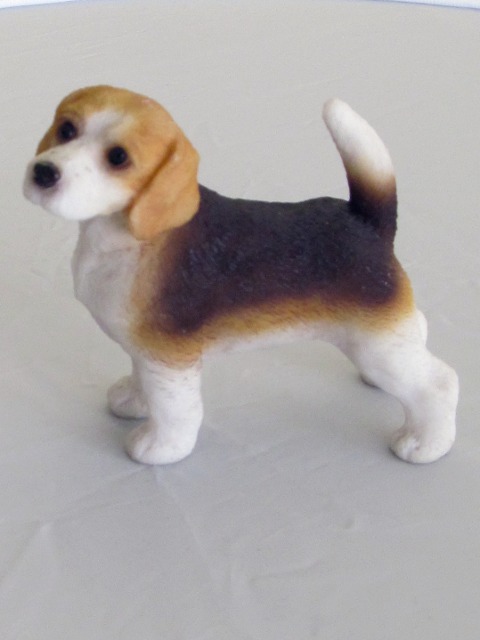}\\
\includegraphics[height=\myfigheight]{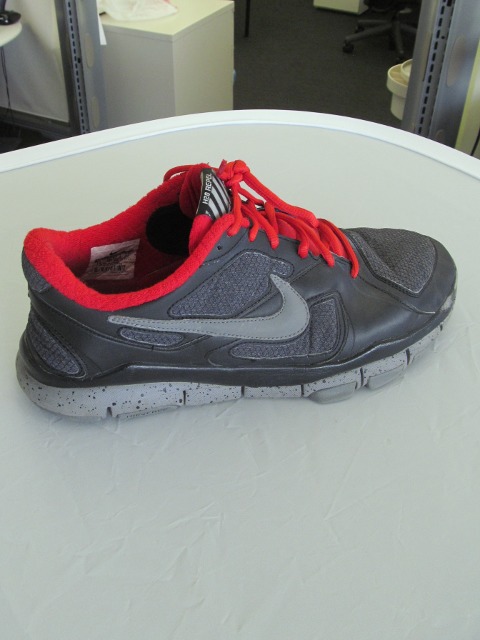}\\
\includegraphics[height=\myfigheight]{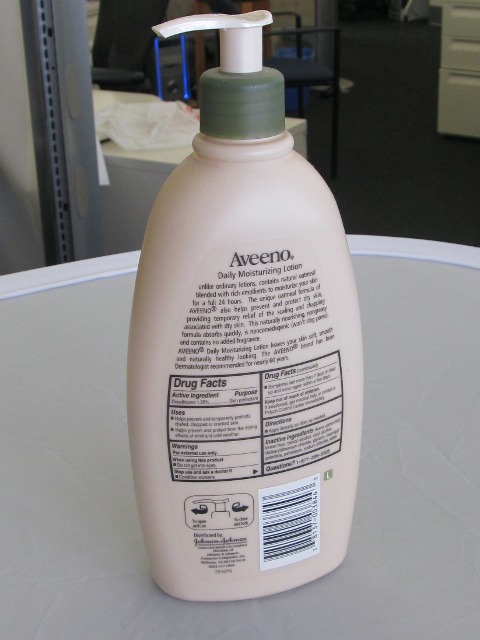}\\
\includegraphics[height=\myfigheight]{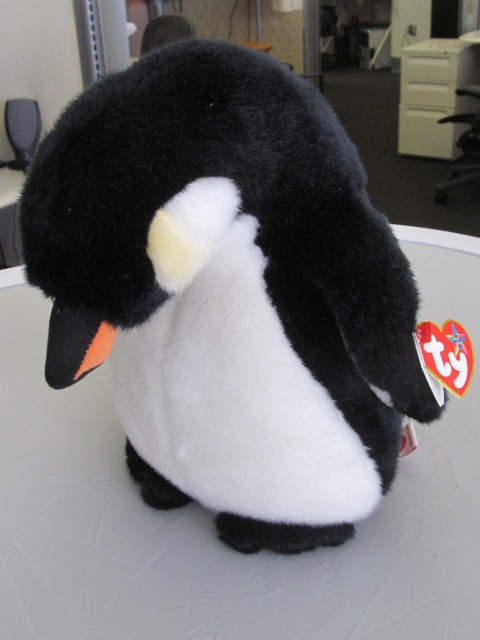}
\end{tabular}}
\subfigure[YAS]{\label{fig:ours}
\begin{tabular}[b]{c}
\includegraphics[height=\myfigheight]{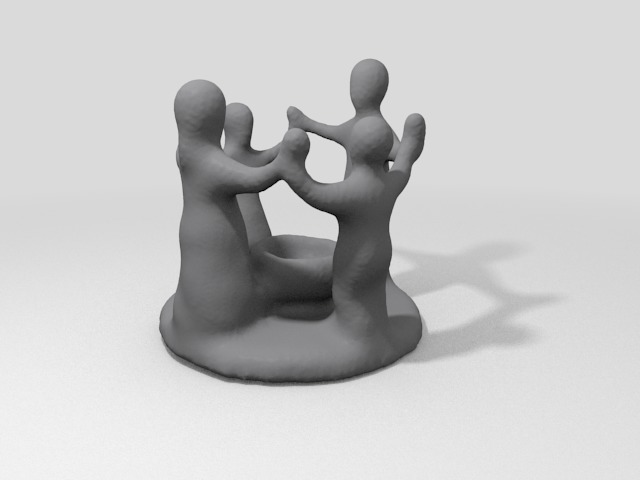}\\
\includegraphics[height=\myfigheight]{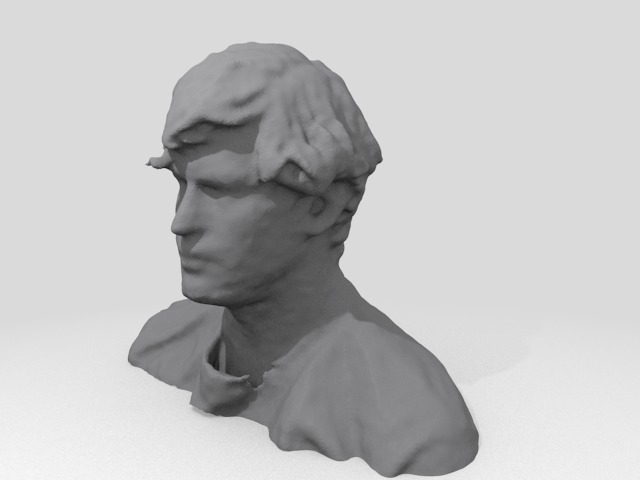}\\
\includegraphics[height=\myfigheight]{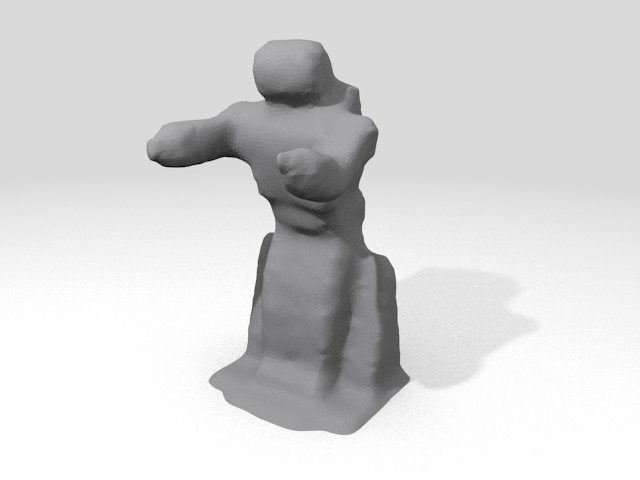}\\
\includegraphics[height=\myfigheight]{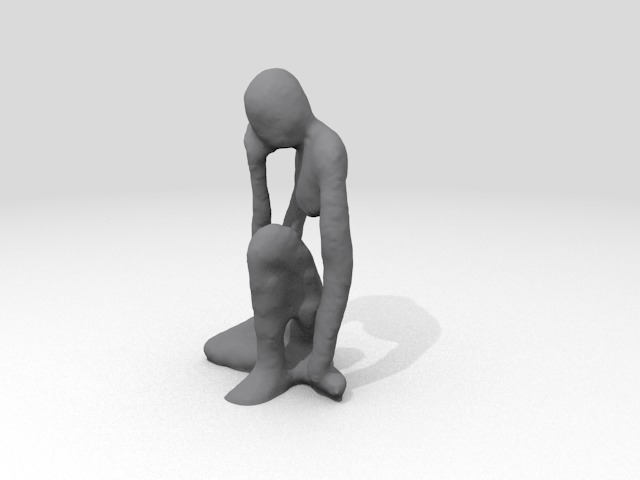}\\
\includegraphics[height=\myfigheight]{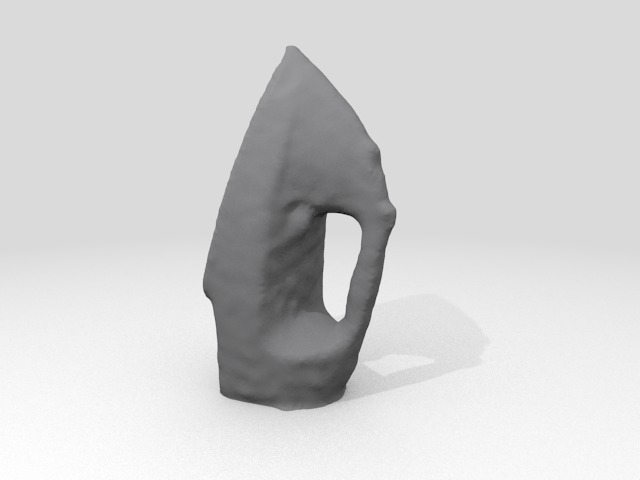}\\
\includegraphics[height=\myfigheight]{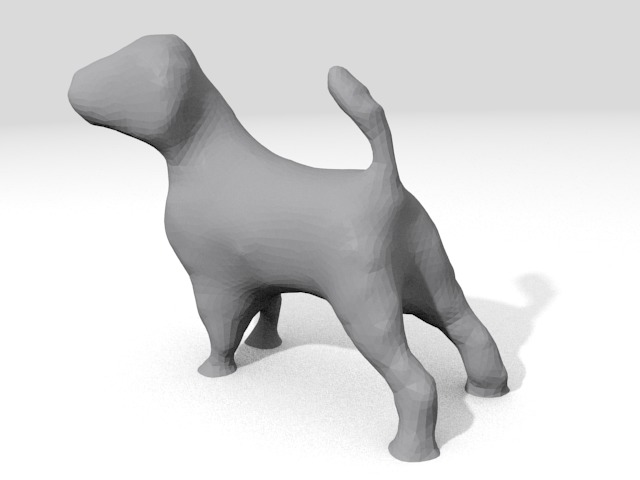}\\
\includegraphics[height=\myfigheight]{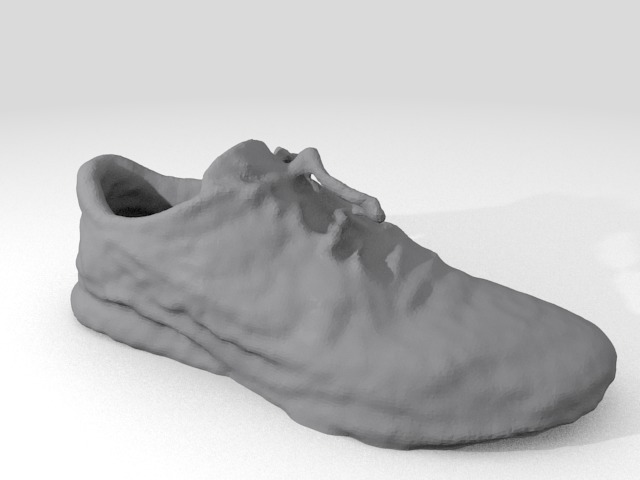}\\
\includegraphics[height=\myfigheight]{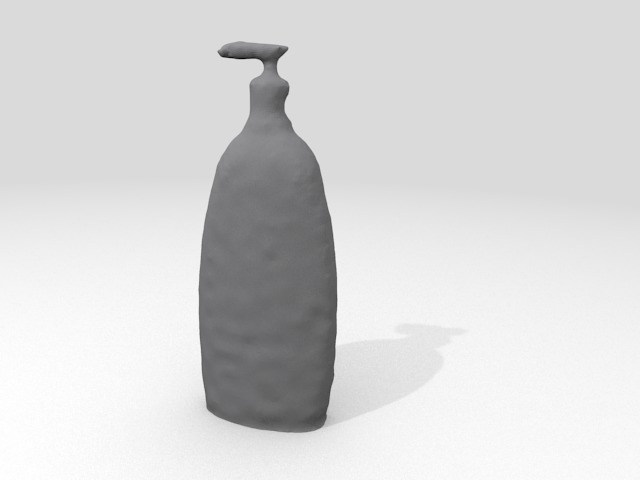}\\
\includegraphics[height=\myfigheight]{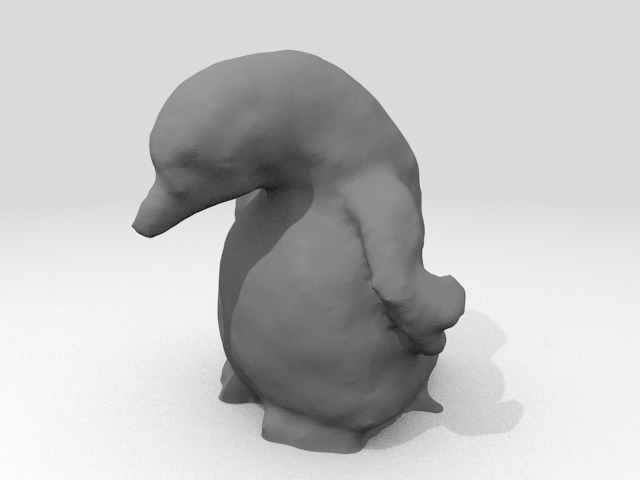}
\end{tabular}}
\subfigure[Scenect]{\label{fig:scenect}
\begin{tabular}[b]{c}
\includegraphics[height=\myfigheight]{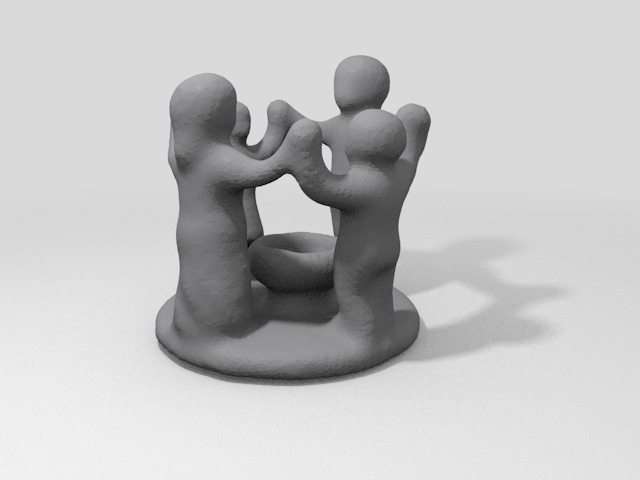}\\
\includegraphics[height=\myfigheight]{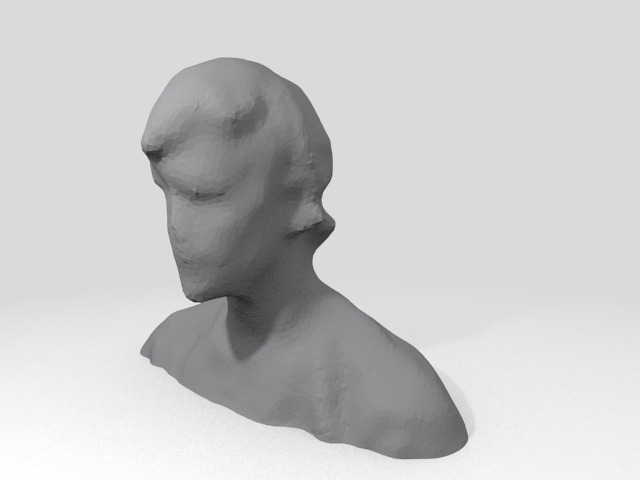}\\
\includegraphics[height=\myfigheight]{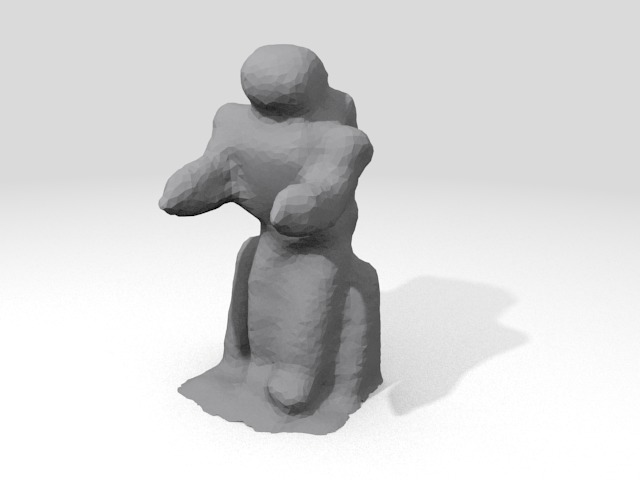}\\
\includegraphics[height=\myfigheight]{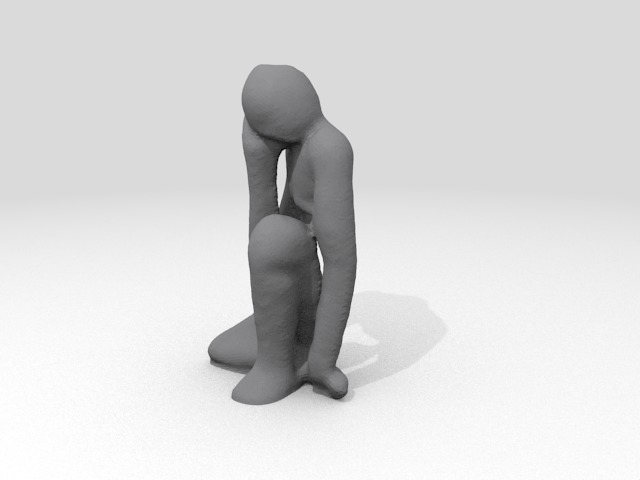}\\
\includegraphics[height=\myfigheight]{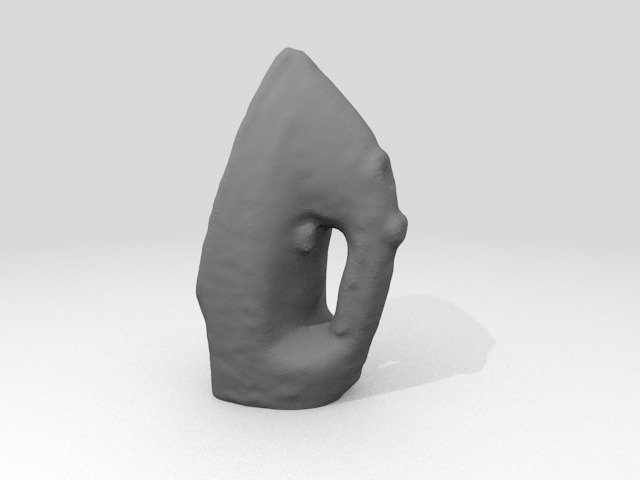}\\
\includegraphics[height=\myfigheight]{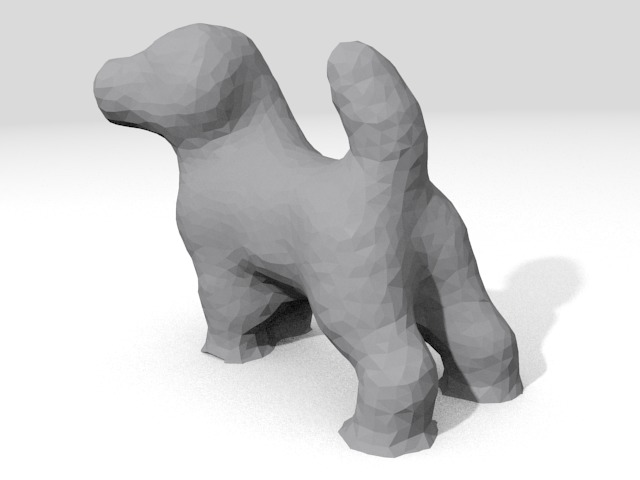}\\
\includegraphics[height=\myfigheight]{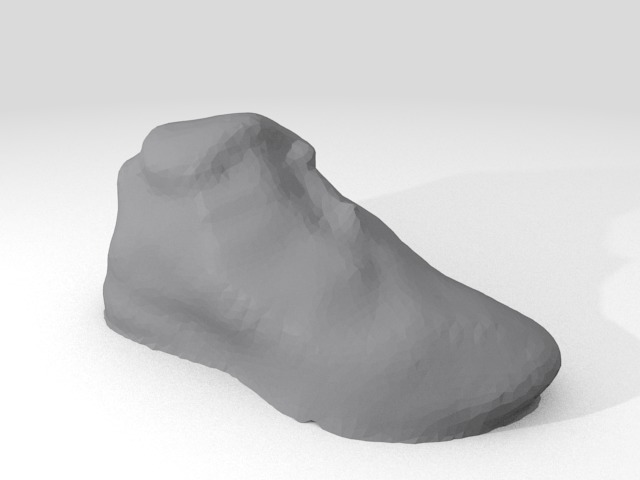}\\
\includegraphics[height=\myfigheight]{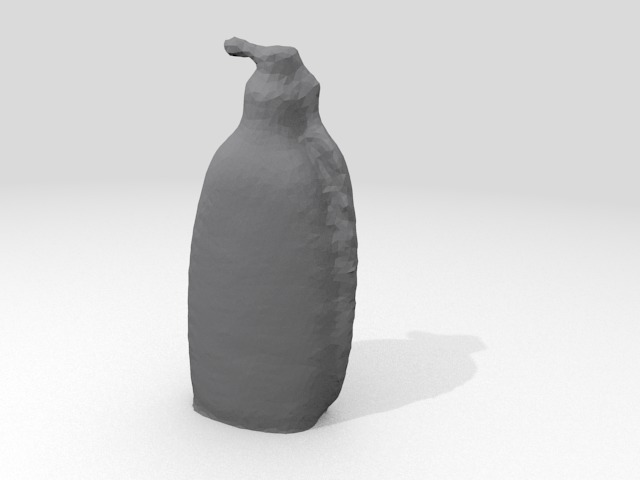}\\
\includegraphics[height=\myfigheight]{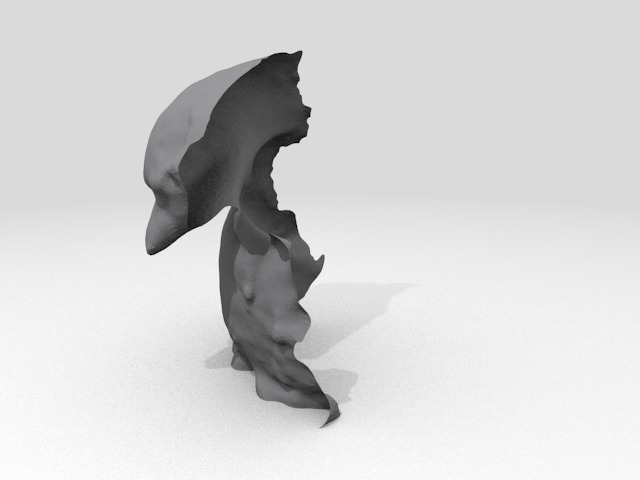}
\end{tabular}}
\subfigure[Kinect Fusion]{\label{fig:kinfu}
\begin{tabular}[b]{c}
\includegraphics[height=\myfigheight]{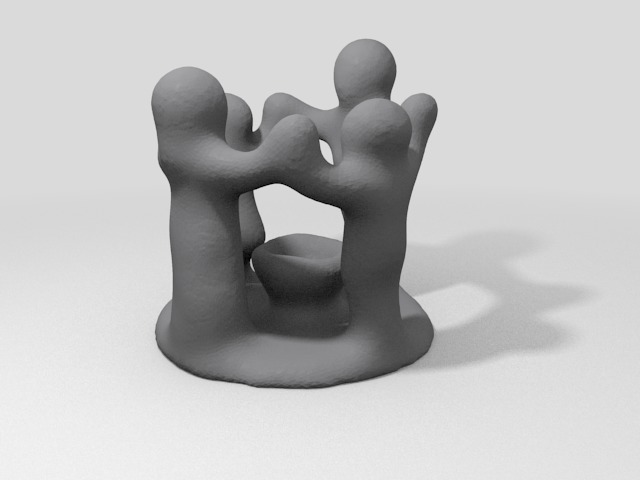}\\
\includegraphics[height=\myfigheight]{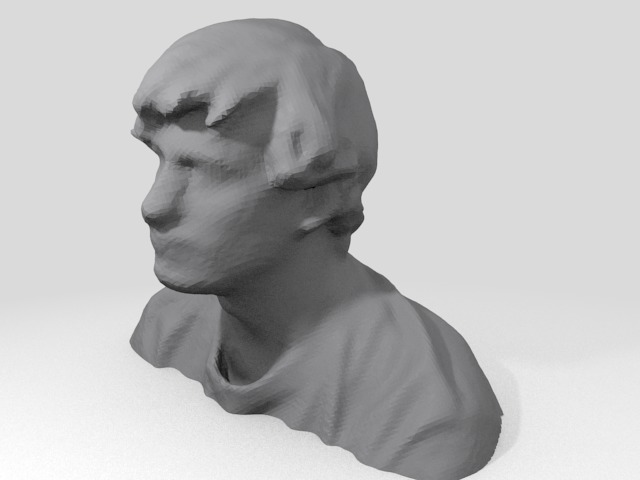}\\
\includegraphics[height=\myfigheight]{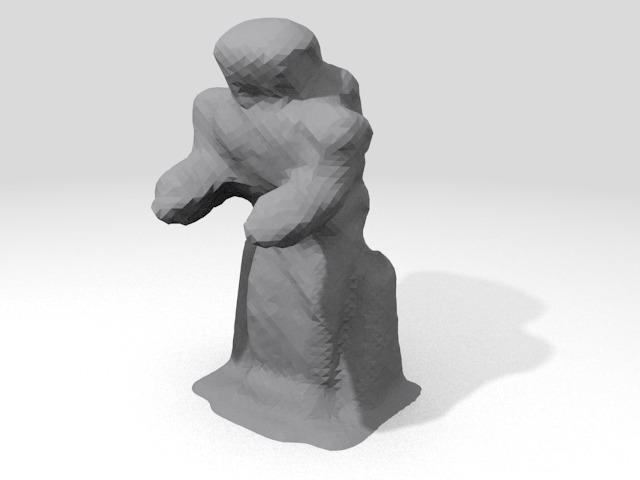}\\
\includegraphics[height=\myfigheight]{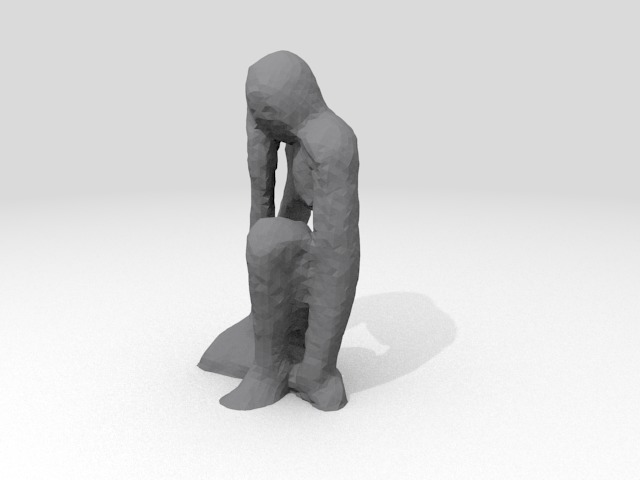}\\
\includegraphics[height=\myfigheight]{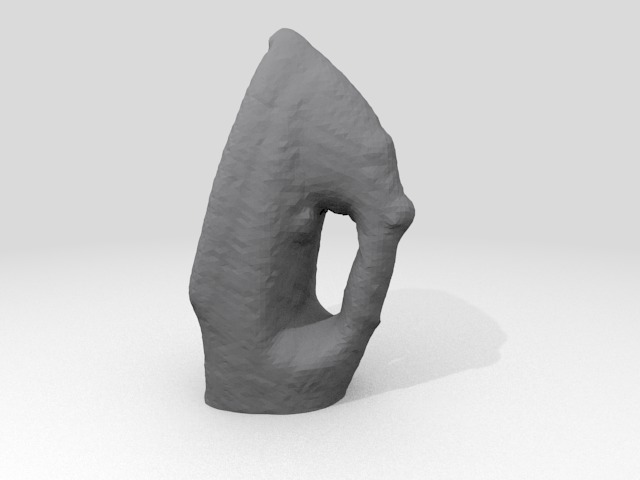}\\
\includegraphics[height=\myfigheight]{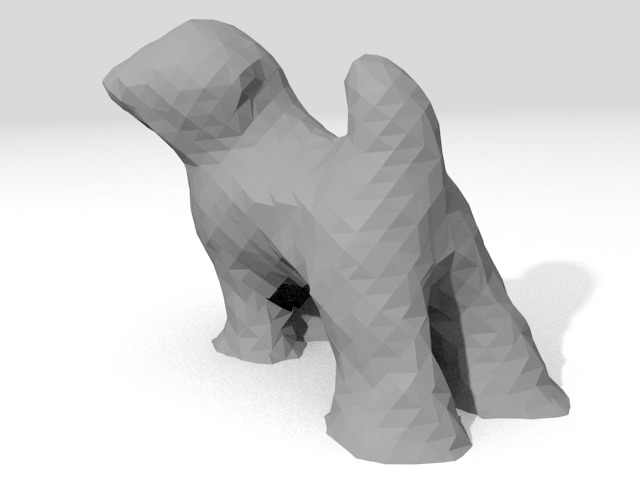}\\
\includegraphics[height=\myfigheight]{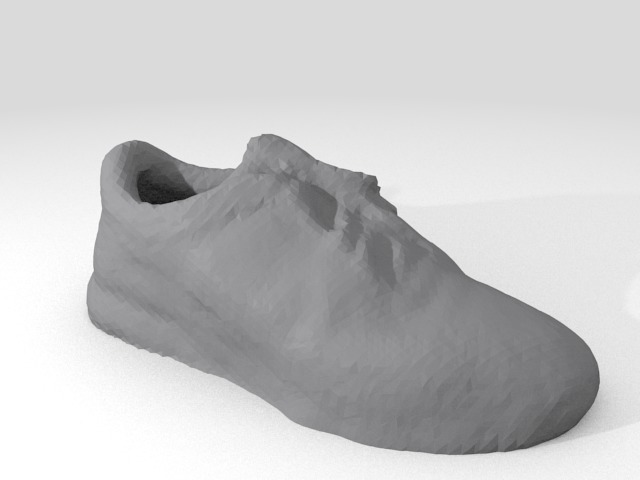}\\
\includegraphics[height=\myfigheight]{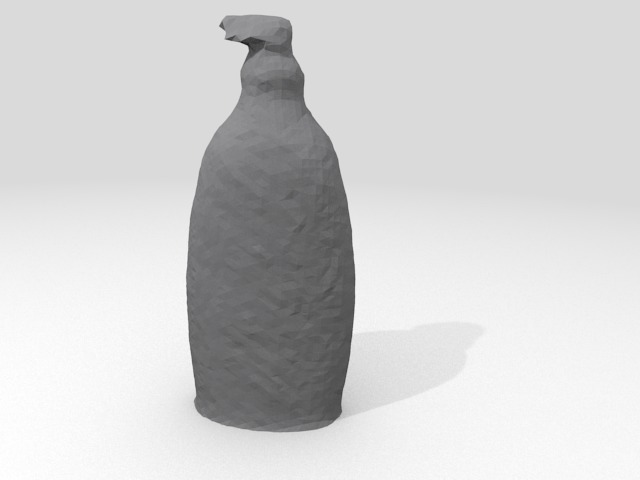}\\
\includegraphics[height=\myfigheight]{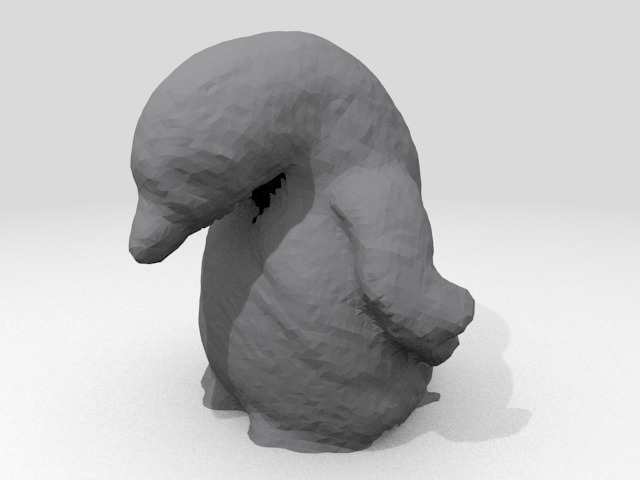}
\end{tabular}}
\caption{Selection of reconstructed free-forms. Let us emphasize that -- to comply with the double-blind review policy -- the bust in the first row is \emph{not} taken from any one of the authors.}\label{fig:examplerec}
\end{figure}

The set of 20 specimens can be divided into two equally-sized groups: The first group contains objects of simple geometry like cubes and cylinders, whose basic dimensions can be measured manually with a tape measure or ruler, see Fig.~\ref{fig:cubes}. Free-forms of sizes ranging from just a few centimeters (fifth row of Fig.~\ref{fig:examplerec}) to the size of a human upper body (first row of Fig.~\ref{fig:examplerec}) make up the second group. The ``ground truth'' volumes for the first group are listed in the first column of Tab.~\ref{tab:volumes}. Needless to say these are afflicted by their own uncertainty, given that they were determined through measurements with a ruler. Our reconstructions of the cube data set are depicted in Fig.~\ref{fig:cubes}. We obtain the best average relative volume error $(V_{\mathrm{YAS}}-V_{\mathrm{Manual}})/V_{\mathrm{Manual}}$ of $-0.34$ $\%$. The performance of Scenect is comparable, which is somewhat surprising in view of Fig.~\ref{fig:scenect}. The meshes created by the Kinect Fusion explorer are of inferior topological  quality: they contain a high number of non-manifold simplices. This, however, does not seem to affect the volume estimates negatively. Also it can be said that the sensitivity of volumes w.r.t. the ground plane parameters is relatively low. 

As can be seen from the last two columns of Tab.~\ref{tab:volumes}, the Kinect Fusion explorer systematically overestimates the ground truth volume by a significant margin. We conjecture that the issue is rooted in calibration. In fact, an important lesson learned during our experimental studies was that a good calibration can make a difference in error of an order of magnitude. Indeed, Scenect provides a calibration program, but Kinect Fusion Explorer does not. A visualization of Tab.~\ref{tab:volumes} is plotted in Fig.~\ref{fig:plots}. 

Results for the second group of objects are shown in Fig.~\ref{fig:examplerec}. Scenect performs worst among the three compared methods. It must be said, though, that Scenect does not offer mesh reconstruction feature, and the point clouds it exports are not oriented. Normals can be computed by singular value decomposition considering the nearest neighbors of a point, which is probably less accurate than the finite-differences approximations of~\eqref{eq:normal}. The lack of loop-closure whose effect can be better identified in the point cloud in Fig.~\ref{fig:misalign2} carries over to the triangular mesh in row of Fig.~\ref{fig:scenect}. Premature termination of the tracker is responsible for the poor reconstruction of the penguin in the last row of Fig.~\ref{fig:scenect}.

Although it behaved unreliably during volume estimation, Kinect Fusion produced high quality models in real-time, which justifies the tremendous success it had since its inception. Still, the scale bias shows, and a lot of details appear to be missing -- details which are present in our YAS reconstructions despite the fact that the Poisson algorithm is known to possess the characteristics of a low-pass filter (see the discussion in Sect.~\ref{subsec:poisson}). The precision setting in Kinect Fusion is quite rigid since the dimensions of the Cartesian grid have to be fixed \emph{before} the reconstruction process even starts. We found that the maximal resolution of $512\times 512\times 512$ voxels rendered the tracking unstable and/or led to incomplete meshes. 

\subsection{Discussion}

We establish correspondence based on photometry, hence our approach fails in the absence of sufficiently exciting texture. This however does not necessarily need to be on the target object itself but can be found in the background, which may be ``enriched'' since after all, in our application, it is not of interest. The majority of competing approaches, including the two we compare against, can cope with the issue. The main reason is the trade-off between a sparse sampling of viewpoints and aforementioned need for sufficient texture: All previous method implicitly leverage on geometry in the motion estimation stage, which is only made possible by the small baseline between adjacent frames, i.e., through tracking, the disadvantages of which have been discussed in Sect.~\ref{subsec:alignment}. In fact, ICP measures the similarity between two points by their distance, hence endows each of them with a geometry descriptor, although a primitive one, too primitive to support wide-baseline matching. More suitable descriptors are available, but we are not yet considering them here, for already the process of salient point detection let alone the computation of informative geometry descriptors is extremely challenging on noisy, occlusion-ridden, and incomplete depth data such as from RGBD sensors. 

The system has no real-time capabilities. We believe this is not necessary for our target application of small-scale object scanning (unlike e.g. navigation, map-building, or odometry). Considering the low resolution of RGB images, the matching process is a matter of seconds. This bottleneck could be removed by an approximate nearest-neighbors search. Our impression is that our system requires the same or even less overall acquisition time compared to e.g. Kinect Fusion, which requires slow and steady motion around the object (and sometimes even a complete reset).

\section{Conclusion}

We described a system for integrating a set of RGBD images of small-scale objects into a geometrically faithful 3-d reconstruction. The system is intended to support researchers in the field of cognitive neuroscience who will use it for acquiring ground truth data for their own experimental studies. To assess the suitability of our 3-d models and those obtained by comparable algorithms, we performed an in-depth analysis of theoretical and empirical kind. There are two main conclusions we would like to draw here: First, the quality of a set of calibration parameters or metric reconstruction can be deceptive. Only quantitative analysis enables veridical information. Second, while the uncertainties of hand-held structured-light scanners may be tremendous in the eyes of the optical metrologist, they help improving studies in the cognitive neurosciences, where manual measurement is still common practice. 

\bibliographystyle{alpha}

\end{document}

%% file: flow.pdf_tex
\begingroup%
  \makeatletter%
  \providecommand\color[2][]{%
    \errmessage{(Inkscape) Color is used for the text in Inkscape, but the package 'color.sty' is not loaded}%
    \renewcommand\color[2][]{}%
  }%
  \providecommand\transparent[1]{%
    \errmessage{(Inkscape) Transparency is used (non-zero) for the text in Inkscape, but the package 'transparent.sty' is not loaded}%
    \renewcommand\transparent[1]{}%
  }%
  \providecommand\rotatebox[2]{#2}%
  \ifx\svgwidth\undefined%
    \setlength{\unitlength}{512bp}%
    \ifx\svgscale\undefined%
      \relax%
    \else%
      \setlength{\unitlength}{\unitlength * \real{\svgscale}}%
    \fi%
  \else%
    \setlength{\unitlength}{\svgwidth}%
  \fi%
  \global\let\svgwidth\undefined%
  \global\let\svgscale\undefined%
  \makeatother%
  \begin{picture}(1,0.75)%
    \put(0,0){\includegraphics[width=\unitlength]{flow.pdf}}%
    \put(0.07403931,0.44697785){\color[rgb]{0,0,0}\makebox(0,0)[lb]{\smash{$z$}}}%
    \put(0.21114683,0.27151909){\color[rgb]{0,0,0}\makebox(0,0)[lb]{\smash{$x$}}}%
    \put(0.1832947,0.38452368){\color[rgb]{0,0,0}\makebox(0,0)[lb]{\smash{$y$}}}%
    \put(0.76434612,0.53985994){\color[rgb]{0,0,0}\makebox(0,0)[lb]{\smash{$\bm{v}$}}}%
    \put(0.40832023,0.54016487){\color[rgb]{0,0,0}\makebox(0,0)[lb]{\smash{$\Gamma$}}}%
  \end{picture}%
\endgroup%

%% file: tr.bbl
\begin{thebibliography}{10}\itemsep=-1pt

\bibitem{Besl1992}
P.~Besl and H.~McKay.
\newblock {A method for registration of 3-D shapes}.
\newblock {\em IEEE T. Pattern Anal.}, 14(2):239--256, 1992.

\bibitem{Cantarella2002}
J.~Cantarella, D.~DeTurck, and H.~Gluck.
\newblock {Vector calculus and the topology of domains in 3-space}.
\newblock {\em Am. Math. Mon.}, 109(5):409--442, 2002.

\bibitem{Charpentier91}
A.~Charpentier.
\newblock {Analyse exp\`{e}rimentale: De quelques elements de la sensation de
  poids}.
\newblock {\em Archieves of Physiology and Normal Pathology}, 3:122--135, 1891.

\bibitem{Ernst02}
M.~Ernst and M.~Banks.
\newblock {Humans integrate visual and haptic information in a statistically
  optimal fashion}.
\newblock {\em Nature}, 415:429--433, 2002.

\bibitem{Fischler1981}
M.~Fischler and R.~Bolles.
\newblock {Random sample consensus: a paradigm for model fitting with
  applications to image analysis and automated cartography}.
\newblock {\em Commun. ACM}, 24(6):381--395, 1981.

\bibitem{Flanagan00}
J.~Flanagan and M.~Beltzner.
\newblock {Independence of perceptual and sensorimotor predictions in the
  size-weight illusion}.
\newblock {\em Nature Neurosci.}, 3:737--741, 2000.

\bibitem{Frayman81}
B.~Frayman and W.~Dawson.
\newblock {The effect of object shape and mode of presentation on judgments of
  apparent volume}.
\newblock {\em Percept. Psychophys.}, 29:56--62, 1981.

\bibitem{Goodale11}
M.~Goodale.
\newblock {Transforming vision into action}.
\newblock {\em Vision Res.}, 51(13):1567--1587, 2011.

\bibitem{Grandy06}
M.~Grandy and D.~Westwood.
\newblock {Opposite perceptual and sensorimotor responses to a size-weight
  illusion}.
\newblock {\em J. Neurophysiol.}, 95:3887--3892, 2006.

\bibitem{Henry2012}
P.~Henry, M.~Krainin, E.~Herbst, X.~Ren, and D.~Fox.
\newblock {RGB-D mapping: Using Kinect-style depth cameras for dense 3D
  modeling of indoor environments}.
\newblock {\em Int. J. Robot. Res.}, 31(5):647--663, 2012.

\bibitem{DanielHerreraC2012}
D.~Herrera, J.~Kannala, and J.~Heikkil\"{a}.
\newblock {Joint depth and color camera calibration with distortion
  correction.}
\newblock {\em IEEE T. Pattern Anal.}, 34(10):2058--64, 2012.

\bibitem{Kazhdan2006}
M.~Kazhdan, M.~Bolitho, and H.~Hoppe.
\newblock {Poisson Surface Reconstruction}.
\newblock {\em Eurographics SGP}, 1:61--70, 2006.

\bibitem{Kazhdan2013}
M.~Kazhdan and H.~Hoppe.
\newblock Screened poisson surface reconstruction.
\newblock {\em ACM Trans. Graph}, 32(1):1--13, 2013.

\bibitem{Lorensen1987}
W.~Lorensen and H.~Cline.
\newblock {Marching cubes: A high resolution 3D surface construction
  algorithm}.
\newblock {\em ACM SIGGRAPH '87}, 21(4):163--169, 1987.

\bibitem{Lowe1999}
D.~Lowe.
\newblock {Object recognition from local scale-invariant features}.
\newblock {\em Proc. ICCV IEEE}, 1:1150--1157, 1999.

\bibitem{Mirtich1996}
B.~Mirtich.
\newblock {Fast and Accurate Computation of Polyhedral Mass Properties}.
\newblock {\em Journal of Graphics Tools}, 1(2):31--50, 1996.

\bibitem{Newcombe2011}
R.~Newcombe, D.~Molyneaux, D.~Kim, P.~Koli, A.~Davison, J.~Shotton, S.~Hodges,
  and A.~Fitzgibbon.
\newblock {KinectFusion: Real-Time Dense Surface Mapping and Tracking}.
\newblock {\em IEEE Proc. ISMAR}, 1:127--136, 2011.

\bibitem{Rock64}
I.~Rock and J.~Victor.
\newblock {Vision and Touch: An Experimentally Created Conflict between the Two
  Senses}.
\newblock {\em Science}, 143(3606):594--596, 1964.

\bibitem{Tam2013}
G.~K.~L. Tam, Z.-Q. Cheng, Y.-K. Lai, F.~C. Langbein, Y.~Liu, D.~Marshall,
  R.~R. Martin, X.-F. Sun, and P.~L. Rosin.
\newblock {Registration of 3D point clouds and meshes: a survey from rigid to
  nonrigid.}
\newblock {\em IEEE T. Vis. Comput. Gr.}, 19(7):1199--217, 2013.

\bibitem{Williams2001}
J.~Williams and M.~Bennamoun.
\newblock Simultaneous registration of multiple corresponding point sets.
\newblock {\em Comput. Vis. Image Und.}, 81(1):117--142, 2001.

\end{thebibliography}
